\pdfoutput=1
\documentclass[conference,a4paper]{IEEEtran}
\IEEEoverridecommandlockouts
\usepackage{amsmath}
\usepackage{newtxtext,newtxmath} 
\usepackage{cite}
\usepackage{graphicx}
\usepackage{booktabs}
\usepackage{multirow}
\usepackage{array}
\usepackage{tabularx}
\usepackage{xcolor}
\usepackage{url}
\usepackage[hidelinks]{hyperref}
\usepackage{eso-pic} 
\graphicspath{{figures/}}
\newcommand{\PM}{\ensuremath{\pm}}
\newcolumntype{Y}{>{\raggedright\arraybackslash}X}
\renewcommand{\IEEEkeywordsname}{Keywords}
\makeatletter
\def\section{\@startsection{section}{1}{\z@}{2pt plus 2pt minus 1pt}{0.1pt plus 1pt}{\normalfont\normalsize\centering\scshape}}
\def\subsection{\@startsection{subsection}{2}{\z@}{0.1pt plus 1pt}{0.1pt plus 1pt}{\normalfont\normalsize\itshape}}
\renewenvironment{abstract}{\par\noindent\fontsize{9}{10.5}\selectfont\hspace*{0.19in}\textit{Abstract}---\ignorespaces}%
  {\par\vspace{4pt}}
\renewenvironment{IEEEkeywords}{\par\noindent\fontsize{9}{10.5}\selectfont\hspace*{0.19in}\textit{Keywords}---\ignorespaces}%
  {\par}
\makeatother
\newcommand{\authorblock}[3]{\IEEEauthorblockN{\fontsize{9}{11}\selectfont #1}%
  \IEEEauthorblockA{\fontsize{9}{11}\selectfont\textit{Artificial Intelligence Technology}\\ \textit{Centre (AITeC)}\\
  \textit{National Centre for Physics (NCP)}\\ Islamabad 44000, Pakistan\\ #2\\ ORCID: #3}}

\begin{document}
\bstctlcite{IEEEexample:BSTcontrol} 

\title{Is It PTSD or the Production? Auditing Shortcut Learning in Video-Based PTSD Detection}

\author{\authorblock{Muhammad Rizwan}{rizwan00277@gmail.com}{0009-0000-9834-9929}
\and \authorblock{Muhammad Awais Khan}{mawaiskhan0486@gmail.com}{0000-0002-9718-1842}
\and \authorblock{Muhammad Ahmad Kamran}{malik.ahmad@ncp.edu.pk}{0000-0002-3942-0059}}

\AddToShipoutPictureBG*{\AtPageLowerLeft{\put(\LenToUnit{0.63in},\LenToUnit{0.35in}){%
  \parbox[b]{7.01in}{\fontsize{7}{8.5}\selectfont \copyright~2026 IEEE. Personal use of this material is permitted. Permission from IEEE must be obtained for all other uses, in any current or future media, including reprinting/republishing this material for advertising or promotional purposes, creating new collective works, for resale or redistribution to servers or lists, or reuse of any copyrighted component of this work in other works. Accepted at ICIC 2026.}}}}
\maketitle

\begin{abstract}
Deep learning models have reported promising results in detecting post-traumatic stress disorder (PTSD) from interview videos. High scores, however, do not show that a model has learned patterns related to the condition. It may instead rely on how videos were recorded, produced or discussed. We investigate this possibility on PTSD in the Wild, a dataset of 634 public videos labeled through self-disclosure. We compare facial, behavioral, vocal and textual models with probes that never see the person. The evaluation becomes progressively stricter: subject-grouped, source-disjoint and metadata-balanced. Under subject-grouped evaluation, container metadata alone reaches an AUC of 0.986, and a background probe with the person removed reaches 0.945, comparable to facial models (0.93--0.96). The frame-selection, attention and temporal components of a reproduced recent system show no consistent advantage over simpler alternatives. On a metadata-balanced subset of 134 videos, facial models reach AUCs of 0.67--0.73, while interviewee-only voice reaches 0.886 and a transcript-only probe 0.923. The text signal persists after diagnostic and military terms are removed, suggesting that broader conversational content contributes. These findings do not establish what the models learn about PTSD itself. They do indicate that source, production and content differences may substantially influence performance on this benchmark.
\end{abstract}


\section{Introduction}
Post-traumatic stress disorder (PTSD) is common after trauma and is often missed, because its diagnosis relies on clinical interviews and self-report. Automatic screening from facial and vocal behavior is therefore an attractive goal for affective computing~\cite{schultebraucks2020,rozgic2014}. Clinical interview corpora are scarce and access-restricted. As a result, PTSD in the Wild~\cite{sawadogo2024}, a set of 634 publicly sourced interview videos, has become the standard benchmark for video-based PTSD detection~\cite{kathan2024,butt2026}. The most recent system, DeepVision~\cite{butt2026}, combines landmark-based ``smart'' frame selection with a ConvNeXt-Tiny backbone~\cite{liu2022convnext} and spatial attention pooling and reports 77.78\% accuracy.

A benchmark assembled from public videos carries a known risk. In PTSD in the Wild, the PTSD videos are mostly veteran testimonials and news clips, while the controls are mostly professionally produced celebrity interviews. A model can then score highly by recognizing \emph{how a video was made} or \emph{what is being discussed} instead of anything about the person. This is \emph{shortcut learning}~\cite{geirhos2020}. It is well documented in medical imaging, where networks learned the hospital or scanner instead of the disease~\cite{zech2018,degrave2021}. Prior work on this benchmark reports accuracy only, on random splits, without any shortcut control.

This paper audits the benchmark rather than proposing a new detector. Its individual diagnostics are established; our contribution is to combine them systematically and apply them to a recent system under controlled conditions. All conclusions concern \emph{benchmark labels} (public self-disclosure), not clinical PTSD. Our contributions are:
\begin{enumerate}
\item Under subject-grouped evaluation, probes that never see a person (metadata, image quality, person-free background) predict the label about as well as face models.
\item Every model is evaluated under four increasingly strict protocols, each time next to a no-person probe on the same folds, with pre-declared tests. Within these protocols we re-evaluate the frame selection, attention and temporal design of DeepVision~\cite{butt2026}.
\item Interviewee voice beats every audio-production control, but a transcript-only probe beats voice, even with diagnosis and military words removed. Conversational content may therefore account for much of the remaining signal.
\end{enumerate}
The rest of the paper covers related work (Section~\ref{sec:related}), data and protocols (Section~\ref{sec:data}), methods (Section~\ref{sec:method}), results (Section~\ref{sec:results}), discussion and limitations (Section~\ref{sec:discussion}), and conclusions (Section~\ref{sec:conclusion}).

\section{Related Work}\label{sec:related}
Facial, vocal and verbal markers have been used to classify PTSD and depression~\cite{schultebraucks2020,rozgic2014}, notably on the clinical DAIC/E-DAIC interview corpora~\cite{ringeval2019}. Reviews warn that recording conditions often differ between patient and control recordings and can confound results~\cite{cummins2015}.

The AVEC challenges established shared audio-visual protocols for depression recognition from clinical interviews~\cite{ringeval2019}. Deep networks have encoded facial appearance and dynamics to estimate depression severity~\cite{zhu2018}, and multi-view speech models have improved PTSD detection in clinical recordings~\cite{zhuang2014}. A systematic review of speech-based psychiatric assessment found small samples and little external validation~\cite{low2020}.

On PTSD in the Wild, Sawadogo et al.~\cite{sawadogo2024} reported audio, visual and text baselines with 3-fold accuracies of 0.88, 0.84 and 0.98 on random splits that are not grouped by subject. Kathan et al.~\cite{kathan2024} reached an in-corpus speech AUC of 0.997, but only 0.70 on a German clinical corpus; this is a cross-corpus test, not an audit of the benchmark. DeepVision~\cite{butt2026} selects MTCNN face crops by raw-landmark cosine change and classifies them with ConvNeXt-T and spatial attention. None of these works uses subject-disjoint splits, tests on unseen sources or controls for shortcuts.

Standard diagnostics for shortcut learning withhold the object of interest or evaluate on held-out sources~\cite{geirhos2020,zech2018}. Mitigations such as domain-adversarial training (DANN)~\cite{ganin2016} and group DRO~\cite{sagawa2020} rarely beat empirical risk minimization (ERM) under real distribution shift~\cite{gulrajani2021}.

\section{Dataset and Evaluation Protocols}\label{sec:data}
PTSD in the Wild~\cite{sawadogo2024} has 634 videos (317 PTSD, 317 control; 89\,h) and provides no subject, source or demographic identifiers. We recovered 13 coarse \emph{sources} with deterministic file-name rules. The PTSD videos come mostly from two veteran testimonial series (189 and 80 videos) and news/TV clips (43). The controls come mostly from GQ (149), WIRED (54) and Vanity Fair (41). Eleven of the 13 sources contain a single label. The two mixed-label sources, TED and ``other'', each contain a single PTSD video. No PTSD video is 480p or 1080p, while 266 of 317 controls are. We linked videos of the same person by names in file names and by FaceNet similarity (threshold 0.70), which gives 580 \emph{subjects}. Face similarity alone recovers only 34\% of the 35 pairs that share a name, so some same-person pairs may remain unlinked. Only 0.02\% of all other pairs exceed the threshold, so false links are rare. In every protocol, no test video has a training or validation video at or above the threshold. All results change by at most 0.009 AUC when the threshold is varied from 0.60 (423 subjects) to 0.90 (603 subjects). All processing followed the dataset license: no video or face leaves local storage, and no figure shows a real face.

We use four evaluation protocols and a source-pair sweep; every video is tested exactly once per protocol.
\begin{itemize}
\item \emph{official3}: the dataset's three random folds ($\approx$211 test videos each). We use them only for comparison with prior work; 74 test videos share a subject with training.
\item \emph{grouped5}: 5-fold StratifiedGroupKFold with subject as the group ($\approx$127 test videos per fold). This is our in-distribution protocol.
\item \emph{sourcedisjoint3}: each test fold pairs PTSD sources with control sources that never appear in its training data. Fold~0 is veteran series~A vs.\ GQ ``10 Things'' (189+136 videos). Fold~1 is series~B vs.\ WIRED and Vanity Fair (80+95). Fold~2 is news/TV and other PTSD clips vs.\ GQ other, TED, talk show and other controls (48+86). Each fold trains on the other two folds, minus any subject that appears in its test fold, so both classes are present in training, but from different sources.
\item \emph{Metadata-balanced fold}: fold~2 of sourcedisjoint3 ($n=134$, 48 PTSD). It was not selected for balance. Its composition follows from the source rules alone: it holds every source outside the two veteran series and the three large control channels, and it was fixed before any model was evaluated on it. We observed afterwards that the metadata probe is at chance on it (AUC 0.505). The fold is balanced in container metadata only; bitrate, face size and speaker count still differ (Section~\ref{sec:voice}).
\item \emph{Source-pair sweep}: to avoid relying on hand-picked folds, the 13 sources are merged into 3 PTSD groups (series~A; series~B; news/TV and other) and 6 control groups (GQ ``10 Things''; GQ other; WIRED; Vanity Fair; TED and talk show; other). Each of the 18 pairs is held out in turn, and training uses all remaining videos minus the test subjects. Both test sources are therefore unseen.
\end{itemize}
Every training portion holds out a subject-grouped inner validation split for early stopping and model selection. No hyper-parameter is tuned on a test fold. Grouped and source-disjoint results are fold-means; metadata-balanced results are the pooled AUC of that single fold.

\section{Method}\label{sec:method}
\begin{figure*}[t]
\centering
\includegraphics[width=0.90\textwidth]{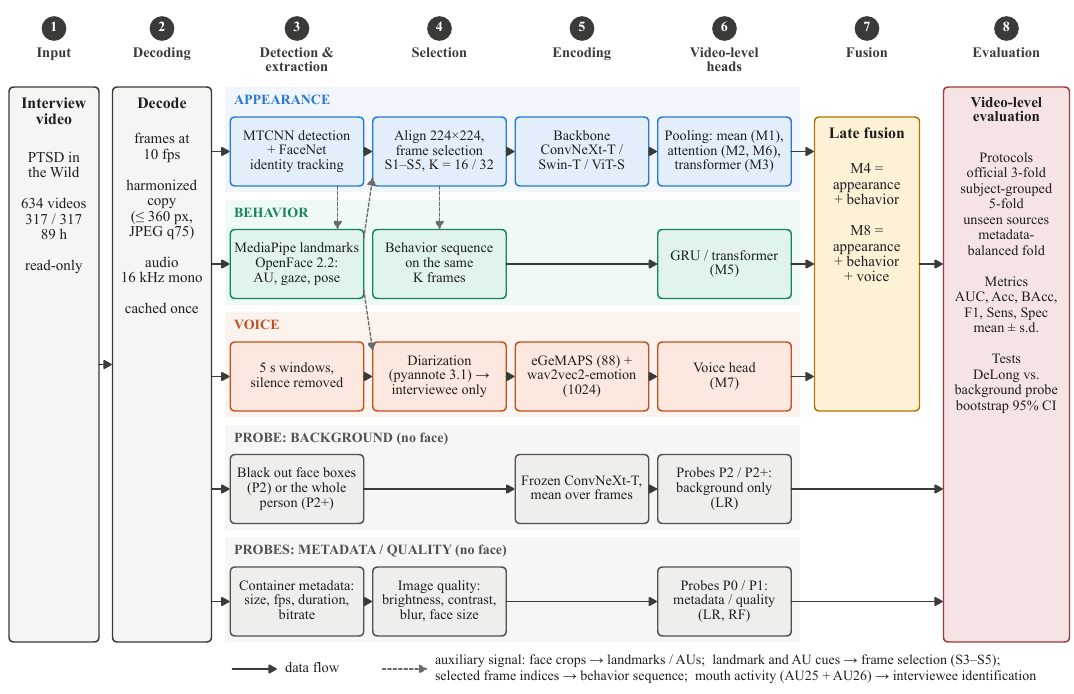}
\caption{Shortcut-aware pipeline. Appearance, behavior and voice streams model the person; two probe lanes never see a face.}
\label{fig:pipeline}
\end{figure*}

Our audit pipeline (Fig.~\ref{fig:pipeline}) processes each video through three streams that model the person (appearance, behavior and voice) and through probe lanes that never see a face. Every model is evaluated against a no-person probe on the same folds, so that a model is credited only with performance beyond what the recording alone predicts. All features are computed once per video and cached (3.2\,M frames; no video failed).

Videos are decoded at 10\,fps. A \emph{harmonized} copy of each frame (at most 360\,px high, JPEG quality 75) removes resolution and compression differences. MTCNN detects faces, a FaceNet identity centroid tracks the interviewee, and faces are aligned and cropped to $224\times224$. We extract frozen ImageNet ConvNeXt-T embeddings, MediaPipe landmarks (with a 68-point Dlib-style subset), and OpenFace~2.2~\cite{baltrusaitis2018} action units (AUs), gaze and head pose. For every 5\,s audio window we compute 88 eGeMAPS functionals and a 1024-d wav2vec2 emotion embedding~\cite{wagner2023}. Speakers are diarized with pyannote.audio~3.1, and speech is transcribed with Whisper small.en.

Each shortcut probe is a video-level logistic regression (LR) or random forest (RF) whose input contains the production but not the person. \textbf{P0}: container metadata (resolution, fps, duration, bitrate). \textbf{P1}: image quality (brightness, contrast, Laplacian blur, face size). \textbf{P2}: the mean embedding of frames with every face box (+25\%) blacked out. \textbf{P2$^{+}$}: the same with the \emph{whole person} removed by MediaPipe selfie segmentation (35\% of the frame on average), so no hair, neck or clothing remains. As a check, MTCNN finds a face in every original frame but in only 0.9\% of person-removed frames (2.1\% on the metadata-balanced fold; 8 frames per video). \textbf{Audio production}: 14 non-person descriptors (speaker count, turn rate, overlap, interviewee talk share, loudness, noise floor, dynamic range). \textbf{Text}: TF-IDF of transcript uni- and bigrams with LR. The vectorizer and classifier are fitted on training-fold transcripts only. This probe is a diagnostic of content and is never used as model input. If a model's AUC is close to a probe's on the same folds, the model shows no benefit beyond the shortcut.

We compare the following models. M1: mean-pooled face embeddings with an MLP. M2: our reproduction of DeepVision's \emph{described method}. No code or curated split was released, so this is not an exact replication of the original experiment. Its $7\times7\times768$ ConvNeXt-T map is pooled by a 2-layer attention scorer with a softmax over the 49 positions, followed by dropout 0.5. It is trained with AdamW (lr $2\times10^{-4}$, weight decay $10^{-2}$), a cosine schedule and horizontal flips on frames chosen by its landmark rule. Video scores are mean frame probabilities. We run both its strict 5-epoch recipe and early stopping. M6 fine-tunes M2 end to end. M3: a temporal transformer (2 layers, 4 heads, CLS token) over $K=32$ frames. M5: a GRU over OpenFace AU, gaze and pose sequences. M7: voice. M8: late fusion of appearance, behavior and voice. Swin-T and ViT-S use the M1 head. As mitigation baselines, the same MLP is also trained with DANN~\cite{ganin2016} and group DRO~\cite{sagawa2020}, using the source as the domain.

Every frame-selection method selects exactly $K\in\{16,32\}$ frames, so any difference comes from \emph{which} frames are chosen. S1 is uniform sampling; S2 is random (3 seeds); S3 is the DeepVision rule (cosine similarity of raw landmark vectors at 2\,fps; frames below the mean similarity, lowest first); S4 applies the same change signal to Procrustes-normalized landmarks; S5 takes the top-$K$ frames by AU-intensity change.

To isolate the interviewee's voice, we select, among the diarized speakers, the one whose segments overlap most with mouth activity (AU25+AU26) of the visible face. We keep only 5\,s windows with at least 80\% of that speaker's speech and at most 10\% of others'.

All metrics are at the video level (AUC, balanced accuracy, F1, sensitivity, specificity), reported as mean\,\PM\,s.d.\ over folds $\times$ 3 seeds (\emph{fold-mean}; seeds 0--2). Paired comparisons use one seed-averaged prediction per video, pooled over the test folds of a protocol (\emph{pooled AUC}, all 634 videos or the 134 of the metadata-balanced fold), so both sides are scored on the same videos. They are tested with DeLong's test~\cite{delong1988} and 2000-sample subject-cluster bootstrap CIs. Equivalence uses two one-sided tests (TOST) with a margin of $\pm0.02$ AUC fixed in advance. We pre-declare three primary hypotheses under Holm control: \textbf{H1} the face model is not better than the background probe; \textbf{H2} interviewee voice beats the audio-production probe on the metadata-balanced fold; \textbf{H3} DeepVision selection is equivalent to uniform sampling.

\section{Results}\label{sec:results}
\begin{table*}[t]
\centering
\caption{Video-level results as the evaluation becomes stricter. official3 ($n=634$, 3 folds), grouped5 ($n=634$, 5 folds) and source-disjoint ($n=634$, 3 folds): fold-mean AUC over folds $\times$ 3 seeds. Metadata-balanced: pooled AUC on fold~2 of sourcedisjoint3 ($n=134$). Margin: AUC minus that of the person-free background P2$^{+}$ (0.681) on the metadata-balanced fold. Last column: official-fold accuracy, for comparison with published numbers. M-rows are trained models; ``probe'' rows are video-level LR or RF classifiers on averaged features. Fig.~\ref{fig:fairer} plots the probe rows for behavior and voice.}
\label{tab:main}
\footnotesize
\setlength{\tabcolsep}{3.5pt}
\begin{tabular}{llcccccc}
\toprule
& & \multicolumn{5}{c}{AUC} & Acc \\
\cmidrule(lr){3-7}\cmidrule(lr){8-8}
Type & Input / model & official3 & grouped5 & source-disj. & meta.-bal. & margin & official3 \\
\midrule
\multirow{4}{*}{No person} & P0 container metadata (RF) & 0.984 & 0.986\PM0.009 & 0.835 & 0.505 & $-$0.176 & -- \\
& P1 image quality, harmonized (RF) & 0.945 & 0.952\PM0.013 & 0.838 & 0.659 & $-$0.022 & -- \\
& P2 background, face boxes out (LR) & 0.954 & 0.954\PM0.012 & 0.832 & 0.597 & $-$0.084 & 88.5\% \\
& P2$^{+}$ background, person removed (LR) & 0.955 & 0.945\PM0.009 & 0.876 & 0.681 & 0 & 87.9\% \\
\midrule
\multirow{6}{*}{Face} & M1 ConvNeXt-T, mean pool & 0.944 & 0.947\PM0.014 & 0.881 & 0.706 & +0.025 & 86.1\% \\
& M2 DeepVision repr.\ (frozen) & 0.937 & 0.933\PM0.013 & 0.840 & 0.687 & +0.006 & 85.4\% \\
& M6 DeepVision repr.\ (fine-tuned) & 0.948 & 0.952\PM0.012 & 0.855 & 0.671 & $-$0.010 & 87.5\% \\
& M3 temporal transformer & 0.940 & 0.933\PM0.019 & 0.862 & 0.667 & $-$0.014 & 85.4\% \\
& Swin-T, mean pool & -- & 0.951\PM0.009 & 0.869 & 0.673 & $-$0.008 & -- \\
& ViT-S, mean pool & -- & 0.961\PM0.012 & 0.893 & 0.725 & +0.044 & -- \\
\midrule
\multirow{2}{*}{Behavior} & M5 OpenFace AU/gaze/pose GRU & -- & 0.869\PM0.029 & 0.785 & 0.643 & $-$0.038 & -- \\
& OpenFace statistics, probe (RF) & 0.917 & 0.918\PM0.011 & 0.853 & 0.690 & +0.009 & -- \\
\midrule
\multirow{4}{*}{Voice} & M7 voice model, all speech & 0.993 & 0.992\PM0.005 & 0.901 & 0.716 & +0.035 & 96.0\% \\
& Voice probe, all speech (LR) & 0.994 & 0.991\PM0.006 & 0.924 & 0.781 & +0.100 & -- \\
& Voice probe, interviewee only (LR) & 0.994 & 0.990 & 0.959 & \textbf{0.886} & \textbf{+0.205} & -- \\
& Prosody only (eGeMAPS), interviewee & -- & -- & 0.773 & 0.648 & $-$0.033 & -- \\
\midrule
Fusion & M8 appearance + behavior + voice & 0.983 & 0.984\PM0.010 & 0.923 & 0.783 & +0.102 & 91.8\% \\
\midrule
Content & Text probe, interviewee transcript & -- & 0.998 & 0.974 & \textbf{0.923} & +0.242 & -- \\
\midrule
\multicolumn{2}{l}{DeepVision as published~\cite{butt2026} (curated subset, accuracy only)} & \multicolumn{5}{c}{--} & 77.78\% \\
\multicolumn{2}{l}{Sawadogo et al., visual~\cite{sawadogo2024} (random 3-fold, accuracy only)} & \multicolumn{5}{c}{--} & 84\% \\
\bottomrule
\end{tabular}
\end{table*}

\subsection{Evidence of a production shortcut}
Under grouped5 (Table~\ref{tab:main}), container metadata alone reaches an AUC of 0.986, and the background with the whole person removed reaches 0.945. Every face model (0.93--0.96), including DeepVision, lies in the same range. Harmonizing resolution and compression changes nothing (background 0.953$\to$0.954), so the shortcut lies in scene, framing and genre rather than image format. For example, the face spans 45\% of the frame height in PTSD videos but 28\% in controls. Every feature family also identifies the source \emph{within} a class: separating the two veteran series (both PTSD) reaches AUC 0.95--0.99 from face, background or voice, and separating GQ from WIRED (both control) reaches 0.98--1.00.

Does the face model rely on this shortcut? In test H1 (grouped5, pooled AUC), the face model M1 differs from the background probe by $-0.007$ AUC (95\% CI $[-0.023, +0.010]$). There is thus no significant face advantage, and any advantage is small (upper bound $+0.010$); formal equivalence within the pre-declared $\pm0.02$ margin is, however, \emph{not} established (TOST $p=0.057$), so we do not claim the two are equal. Face-model and background predictions are strongly correlated (Spearman $\rho=0.83$). Adding face features to the background probe gains only $+0.002$ AUC in-distribution (CI $[-0.007,+0.011]$, $p=0.65$). On unseen sources the gain is $+0.032$ (CI $[+0.012,+0.053]$), so the face carries some information that the background lacks.

\begin{figure*}[t]
\centering
\includegraphics[width=0.84\textwidth]{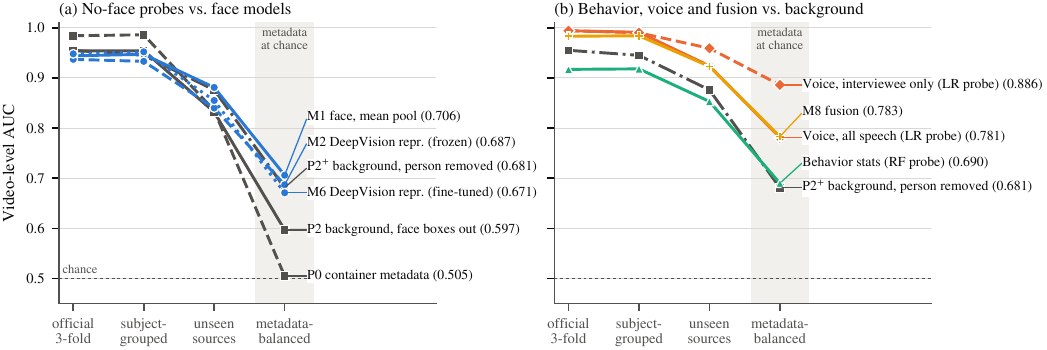}
\caption{AUC as the evaluation becomes stricter, from the leaky official folds to the metadata-balanced fold (shaded), where container metadata is at chance. (a)~Face models, DeepVision included, fall together with the no-face probes. (b)~Voice keeps a margin over the background, which widens when only the interviewee's speech is used. Panel~(b) shows the behavior and voice \emph{probes} of Table~\ref{tab:main}; the trained models M5 and M7 are listed there only.}
\label{fig:fairer}
\end{figure*}

\subsection{Performance under stricter evaluation}
\begin{figure}[!t]
\centering
\includegraphics[width=0.95\columnwidth]{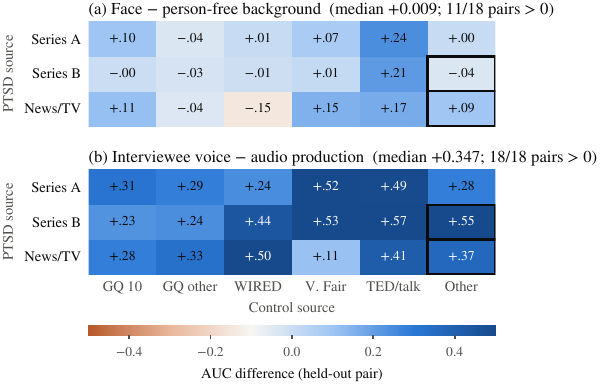}
\caption{Source-pair sweep. Each cell holds out one PTSD and one control source group, both unseen in training. (a)~Face AUC minus person-free background AUC. (b)~Interviewee-voice AUC minus audio-production AUC. Boxed cells: pairs where container metadata is at chance. The TED/talk group has only 9 control videos.}
\label{fig:sweep}
\end{figure}

Fig.~\ref{fig:fairer} follows each input across the protocols; the two strict settings answer different questions. On unseen sources (sourcedisjoint3), every model degrades but remains high (face models 0.84--0.89; appearance 0.947$\to$0.881, DeepVision 0.933$\to$0.840). The person-free background is equally high (0.876), because new channels still differ from each other in genre and framing. The metadata-balanced fold removes container differences as well, and there face models (0.67--0.73) are close to the background (0.681). To check that these results do not depend on hand-picked folds, Fig.~\ref{fig:sweep} repeats the test for all 18 source pairs, each with both sources unseen in training. Metadata separates the classes perfectly in 9 pairs and is at chance in only 2 (boxed cells). In Fig.~\ref{fig:sweep}a most cells are close to zero: the face beats the person-free background in 11 of 18 pairs, by a median of only $+0.009$ AUC. The largest gains ($+0.17$ to $+0.24$) occur against the TED/talk group, which has only 9 control videos. In Fig.~\ref{fig:sweep}b every cell is clearly positive: interviewee voice exceeds the audio-production probe in all 18 pairs (Section~\ref{sec:voice}).

\subsection{Re-evaluating DeepVision's design claims}
On the official folds our reproduction of DeepVision reaches 85.4\% accuracy (frozen) and 87.5\% (fine-tuned); the strict 5-epoch recipe gives 84.9\% and 87.3\%. The method therefore performs well when re-implemented, above the published 77.78\%, which was obtained on a curated subset that is not available. However, on the same folds the background probe reaches 88.5\% with only the face boxes removed and 87.9\% with the whole person removed. Accuracy at this level is thus not, by itself, evidence of PTSD detection.

For frame selection, at an equal budget (Table~\ref{tab:selection}), DeepVision selection and uniform sampling are \emph{equivalent} (H3: difference $+0.001$, 90\% CI $[-0.006, +0.007]$, Holm-adjusted TOST $p<10^{-4}$), and so are all frames at 2\,fps and at 10\,fps. The result is the same with the order-sensitive transformer head (S1--S3: 0.932--0.937). On unseen sources, DeepVision selection is slightly \emph{below} uniform sampling (0.859 vs.\ 0.874 at $K=16$). Our raw-landmark cosine stays close to 1 (mean 0.9987, compared with the 0.9627 reported in~\cite{butt2026}). Without released code, this gap most likely reflects unreported details of landmark normalization or face-crop alignment. The conclusion holds regardless, at both budgets, both frame rates and with both heads. The variation of the raw cosine follows head and face-box motion (per-video $\rho=0.52$) more than expression change ($\rho=0.36$).

\begin{figure}[!t]
\centering
\includegraphics[width=0.86\columnwidth]{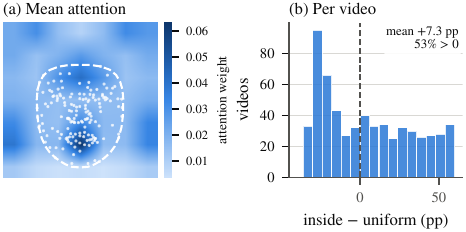}
\caption{Spatial attention of the DeepVision reproduction (M2), drawn on a mean landmark template rather than a real face. (a)~Mean attention over all videos; dashed line: face outline. (b)~Per video, attention inside the face minus the share expected under uniform attention (percentage points).}
\label{fig:attention}
\end{figure}

For spatial attention, only 38.7\% of M2's attention mass falls inside each video's own face outline, compared with 31.5\% expected under uniform attention. Only 53\% of videos place more attention on the face than uniform attention would (Fig.~\ref{fig:attention}), and many attend mainly to hair, clothing and crop borders. The temporal transformer is \emph{worse} than mean pooling (pooled AUC $-0.017$, CI $[-0.028,-0.006]$, $p=0.001$). Fine-tuning helps in-distribution (M6 vs.\ M2, pooled AUC: 0.957 vs.\ 0.940, $p=0.0002$) but not on unseen sources (0.855 vs.\ 0.840).

As mitigation baselines, DANN and group DRO change AUC by at most 0.006 on unseen sources and 0.014 on the metadata-balanced fold relative to ERM. With 11 of 13 sources single-label, source-invariant features are largely label-invariant.

\begin{table}[t]
\centering
\caption{Frame selection at a fixed budget of $K$ frames per video (fold-mean AUC). Mean pooling + LR on frozen ConvNeXt-T embeddings, and the temporal transformer M3 ($K=32$).}
\label{tab:selection}
\footnotesize
\setlength{\tabcolsep}{2.2pt}
\begin{tabular}{lcccccc}
\toprule
& \multicolumn{2}{c}{Mean, grp5} & \multicolumn{2}{c}{Mean, src-dj.} & \multicolumn{2}{c}{M3, $K{=}32$} \\
\cmidrule(lr){2-3}\cmidrule(lr){4-5}\cmidrule(lr){6-7}
Method & $K{=}16$ & $K{=}32$ & $K{=}16$ & $K{=}32$ & grp5 & src-dj. \\
\midrule
All frames, 10\,fps & \multicolumn{2}{c}{0.940} & \multicolumn{2}{c}{0.873} & -- & -- \\
S1 uniform & 0.940 & 0.940 & 0.874 & 0.870 & 0.933 & 0.862 \\
S2 random & 0.937 & 0.937 & 0.877 & 0.873 & 0.932 & 0.858 \\
S3 DeepVision & 0.939 & 0.942 & 0.859 & 0.863 & 0.937 & 0.861 \\
S4 normalized lmk. & 0.940 & 0.944 & 0.865 & 0.869 & 0.937 & 0.857 \\
S5 AU change & 0.934 & 0.933 & 0.859 & 0.853 & 0.922 & 0.841 \\
\bottomrule
\end{tabular}
\end{table}

\subsection{What survives: voice, or what is said?}\label{sec:voice}
The metadata-balanced fold is not fully balanced. PTSD videos there have higher bitrate (standardized mean difference, SMD 0.67), larger faces (SMD 0.70) and more diarized speakers (3.8 vs.\ 1.9; SMD 0.88). Loudness, noise floor and dynamic range are balanced. Voice is the strongest stream on this fold, and we tested three alternative explanations.

\emph{Other speakers.} For the same LR voice probe (Table~\ref{tab:main}), restricting the input to the interviewee \emph{raises} its AUC from 0.781 to 0.886 (gain $+0.105$, CI $[+0.037,+0.167]$). \emph{Audio production.} A probe trained on all 14 audio-production descriptors reaches only 0.481, and interviewee voice beats it by $+0.405$ (CI $[+0.291,+0.516]$; H2, Holm-adjusted $p<10^{-4}$). Interviewee voice also beats the audio-production probe in all 18 source pairs (median $+0.347$; Fig.~\ref{fig:sweep}b). Computed only over PTSD--control pairs with the \emph{same number of speakers}, its AUC remains 0.893, against 0.59 for the background. It also holds in videos with at most two speakers (0.870; 96 videos, 18 PTSD) and in videos where the interviewee speaks at least 80\% of the time (0.978; 85 videos, 12 PTSD). The interviewee selection itself was not verified by listening; these stratified results are indirect evidence that it works. \emph{Content.} Prosody alone (eGeMAPS) gives only 0.648, so most of the signal comes from the self-supervised embedding, which also encodes phonetic and lexical information. A text-only probe on the transcripts does better than voice (0.923). Its most PTSD-indicative terms name the label or its context (\emph{PTSD}, \emph{VA}, \emph{veterans}, \emph{therapy}), and its most control-indicative terms come from lifestyle videos (\emph{favorite}, \emph{essentials}). Explicit disclosure is not required, however. When we delete diagnosis words (\emph{PTSD}, \emph{trauma}, \emph{disorder}, \emph{diagnosis}, \emph{stress}) before training, the AUC remains 0.927. When we also delete military, veteran, treatment and trauma-type words (about 40 patterns), it is still 0.933. This is consistent with broader conversational content and genre, rather than a few label words. Voice and text predictions are correlated ($\rho=0.71$), and fusing them does not improve on text alone. After the text prediction is regressed out, the voice ranking keeps an AUC of 0.662. The results suggest that conversational content may account for a substantial part of the voice signal, while a modest part is not explained by the words.

\section{Discussion and Limitations}\label{sec:discussion}
Our experiments show that person-free probes come within 0.01 AUC of face models in-distribution, and that face models fall to 0.67--0.73 on the metadata-balanced fold while voice and text remain higher. We \emph{interpret} this as indicating that benchmark performance may partly reflect source and production recognition, and that conversational content and genre may drive much of the voice and text signal. These explanations are plausible, but the data cannot prove them. We recommend that future work report (i)~subject-grouped and source-disjoint results; (ii)~metadata, person-free background, audio-production and text references on the same folds; and (iii)~paired AUC differences with subject-cluster CIs. Self-disclosed labels, uncertain source associations, automatic subject linking and interviewee selection, the small metadata-balanced fold and dataset confounds limit generalization. Validation on independently sourced data is required before real-world deployment.

\section{Conclusion}\label{sec:conclusion}
We examined whether high performance on PTSD in the Wild reflects PTSD-related patterns or differences in production and content. Person-free probes performed comparably to facial models, and on a metadata-balanced subset facial models declined while voice and transcript models remained more predictive. The tested frame-selection, attention and temporal components showed no consistent advantage. These findings do not establish that facial or vocal behavior carries no PTSD-relevant information. They suggest that such information is difficult to separate from source, production and content effects in this benchmark. Resolving this needs clinically labeled data in which PTSD status varies across sources and topics. Until then, high benchmark scores alone should not be taken as evidence of clinically reliable PTSD detection.

\section*{Acknowledgment}
Claude (Anthropic) was used for language editing, proofreading, \LaTeX{} formatting, and assistance with code implementation. All AI-assisted outputs were reviewed and verified by the authors, who take full responsibility for the manuscript, methodology, and results.

\bibliographystyle{IEEEtran}
\bibliography{refs}

@article{sawadogo2024,
  author  = {Sawadogo, Moctar Abdoul Latif and Pala, Furkan and Singh, Gurkirat and Selmi, Imen and Puteaux, Pauline and Othmani, Alice},
  title   = {{PTSD} in the wild: a video database for studying post-traumatic stress disorder recognition in unconstrained environments},
  journal = {Multimedia Tools Appl.},
  volume  = {83},
  number  = {14},
  pages   = {42861--42883},
  year    = {2024},
  note    = {arXiv:2209.14085},
  doi     = {10.1007/s11042-023-17203-x}
}

@inproceedings{butt2026,
  author    = {Butt, Abdul Mateen and Asghar, Ismaeel and Razzaq, Fiza Abdul},
  title     = {{DeepVision}: Video-Based {PTSD} Detection and Decision Support System},
  booktitle = {Proc. 7th Int. Conf. on Advancements in Computational Sciences (ICACS)},
  year      = {2026},
  doi       = {10.1109/ICACS69208.2026.11433160},
  pages     = {1--6}
}

@inproceedings{kathan2024,
  author    = {Kathan, Alexander and B{\"u}rger, Martin and Triantafyllopoulos, Andreas and Milkus, Sabrina and Hohmann, Jonas and Muderlak, Pauline and Schottdorf, J{\"u}rgen and Musil, Richard and Schuller, Bj{\"o}rn and Amiriparian, Shahin},
  title     = {Real-world {PTSD} Recognition: A Cross-corpus and Cross-linguistic Evaluation},
  booktitle = {Proc. Interspeech},
  pages     = {487--491},
  year      = {2024},
  doi       = {10.21437/Interspeech.2024-493}
}

@article{schultebraucks2020,
  author  = {Schultebraucks, Katharina and Yadav, Vijay and Shalev, Arieh Y. and Bonanno, George A. and Galatzer-Levy, Isaac R.},
  title   = {Deep learning-based classification of posttraumatic stress disorder and depression following trauma utilizing visual and auditory markers of arousal and mood},
  journal = {Psychol. Med.},
  volume  = {52},
  number  = {5},
  pages   = {957--967},
  year    = {2022},
  doi     = {10.1017/S0033291720002718}
}

@inproceedings{rozgic2014,
  author    = {Rozgi{\'c}, Viktor and Vazquez-Reina, Amelio and Crystal, Michael and Srivastava, Amit and Tan, Veasna and Berka, Chris},
  title     = {Multi-modal prediction of {PTSD} and stress indicators},
  booktitle = {Proc. IEEE ICASSP},
  year      = {2014},
  pages     = {3636--3640},
  doi       = {10.1109/ICASSP.2014.6854279}
}

@inproceedings{zhuang2014,
  author    = {Zhuang, Xiaodan and Rozgi{\'c}, Viktor and Crystal, Michael and Marx, Brian P.},
  title     = {Improving speech-based {PTSD} detection via multi-view learning},
  booktitle = {Proc. IEEE Spoken Language Technology Workshop (SLT)},
  pages     = {260--265},
  year      = {2014},
  doi       = {10.1109/SLT.2014.7078584}
}

@inproceedings{ringeval2019,
  author    = {Ringeval, Fabien and others},
  title     = {{AVEC} 2019 Workshop and Challenge: State-of-Mind, Detecting Depression with {AI}, and Cross-Cultural Affect Recognition},
  booktitle = {Proc. 9th Int. Audio/Visual Emotion Challenge and Workshop (AVEC)},
  pages     = {3--12},
  year      = {2019},
  doi       = {10.1145/3347320.3357688}
}

@article{geirhos2020,
  author  = {Geirhos, Robert and Jacobsen, J{\"o}rn-Henrik and Michaelis, Claudio and Zemel, Richard and Brendel, Wieland and Bethge, Matthias and Wichmann, Felix A.},
  title   = {Shortcut learning in deep neural networks},
  journal = {Nature Mach. Intell.},
  volume  = {2},
  number  = {11},
  pages   = {665--673},
  year    = {2020},
  doi     = {10.1038/s42256-020-00257-z}
}

@article{zech2018,
  author  = {Zech, John R. and Badgeley, Marcus A. and Liu, Manway and Costa, Anthony B. and Titano, Joseph J. and Oermann, Eric Karl},
  title   = {Variable generalization performance of a deep learning model to detect pneumonia in chest radiographs: A cross-sectional study},
  journal = {PLoS Med.},
  volume  = {15},
  number  = {11},
  pages   = {e1002683},
  year    = {2018},
  doi     = {10.1371/journal.pmed.1002683}
}

@article{degrave2021,
  author  = {DeGrave, Alex J. and Janizek, Joseph D. and Lee, Su-In},
  title   = {{AI} for radiographic {COVID-19} detection selects shortcuts over signal},
  journal = {Nature Mach. Intell.},
  volume  = {3},
  pages   = {610--619},
  year    = {2021},
  number  = {7},
  doi     = {10.1038/s42256-021-00338-7}
}

@inproceedings{liu2022convnext,
  author    = {Liu, Zhuang and Mao, Hanzi and Wu, Chao-Yuan and Feichtenhofer, Christoph and Darrell, Trevor and Xie, Saining},
  title     = {A {ConvNet} for the 2020s},
  booktitle = {Proc. IEEE/CVF CVPR},
  pages     = {11966--11976},
  year      = {2022},
  doi       = {10.1109/CVPR52688.2022.01167}
}

@inproceedings{baltrusaitis2018,
  author    = {Baltru{\v{s}}aitis, Tadas and Zadeh, Amir and Lim, Yao Chong and Morency, Louis-Philippe},
  title     = {{OpenFace} 2.0: Facial Behavior Analysis Toolkit},
  booktitle = {Proc. IEEE Int. Conf. on Automatic Face \& Gesture Recognition (FG)},
  pages     = {59--66},
  year      = {2018},
  doi       = {10.1109/FG.2018.00019}
}

@article{wagner2023,
  author  = {Wagner, Johannes and Triantafyllopoulos, Andreas and Wierstorf, Hagen and Schmitt, Maximilian and Burkhardt, Felix and Eyben, Florian and Schuller, Bj{\"o}rn W.},
  title   = {Dawn of the Transformer Era in Speech Emotion Recognition: Closing the Valence Gap},
  journal = {IEEE Trans. Pattern Anal. Mach. Intell.},
  volume  = {45},
  number  = {9},
  pages   = {10745--10759},
  year    = {2023},
  doi     = {10.1109/TPAMI.2023.3263585}
}

@article{delong1988,
  author  = {DeLong, Elizabeth R. and DeLong, David M. and Clarke-Pearson, Daniel L.},
  title   = {Comparing the areas under two or more correlated receiver operating characteristic curves: a nonparametric approach},
  journal = {Biometrics},
  volume  = {44},
  number  = {3},
  pages   = {837--845},
  year    = {1988},
  doi     = {10.2307/2531595}
}

@article{cummins2015,
  author  = {Cummins, Nicholas and Scherer, Stefan and Krajewski, Jarek and Schnieder, Sebastian and Epps, Julien and Quatieri, Thomas F.},
  title   = {A review of depression and suicide risk assessment using speech analysis},
  journal = {Speech Commun.},
  volume  = {71},
  pages   = {10--49},
  year    = {2015},
  doi     = {10.1016/j.specom.2015.03.004}
}

@article{low2020,
  author  = {Low, Daniel M. and Bentley, Kate H. and Ghosh, Satrajit S.},
  title   = {Automated assessment of psychiatric disorders using speech: A systematic review},
  journal = {Laryngoscope Investig. Otolaryngol.},
  volume  = {5},
  number  = {1},
  pages   = {96--116},
  year    = {2020},
  doi     = {10.1002/lio2.354}
}

@article{ganin2016,
  author  = {Ganin, Yaroslav and Ustinova, Evgeniya and Ajakan, Hana and Germain, Pascal and Larochelle, Hugo and Laviolette, Fran{\c{c}}ois and Marchand, Mario and Lempitsky, Victor},
  title   = {Domain-Adversarial Training of Neural Networks},
  journal = {J. Mach. Learn. Res.},
  volume  = {17},
  number  = {59},
  pages   = {1--35},
  year    = {2016}
}

@inproceedings{sagawa2020,
  author    = {Sagawa, Shiori and Koh, Pang Wei and Hashimoto, Tatsunori B. and Liang, Percy},
  title     = {Distributionally Robust Neural Networks for Group Shifts: On the Importance of Regularization for Worst-Case Generalization},
  booktitle = {Proc. ICLR},
  year      = {2020}
}

@inproceedings{gulrajani2021,
  author    = {Gulrajani, Ishaan and Lopez-Paz, David},
  title     = {In Search of Lost Domain Generalization},
  booktitle = {Proc. ICLR},
  year      = {2021}
}

@article{zhu2018,
  author  = {Zhu, Yu and Shang, Yuanyuan and Shao, Zhuhong and Guo, Guodong},
  title   = {Automated Depression Diagnosis Based on Deep Networks to Encode Facial Appearance and Dynamics},
  journal = {IEEE Trans. Affective Comput.},
  volume  = {9},
  number  = {4},
  pages   = {578--584},
  year    = {2018},
  doi     = {10.1109/TAFFC.2017.2650899}
}

@IEEEtranBSTCTL{IEEEexample:BSTcontrol,
  CTLuse_forced_etal       = "yes",
  CTLmax_names_forced_etal = "6",
  CTLnames_show_etal       = "1"
}

\end{document}